\documentclass[final]{cvpr}

\usepackage{times}
\usepackage{epsfig}
\usepackage{graphicx}
\usepackage{amsmath,amssymb}
\usepackage{mathtools}
\usepackage{algpseudocode}
\usepackage{algorithm}

\usepackage{bm}
\usepackage{multirow}
\usepackage{bbold}
\usepackage{empheq}

\usepackage[pagebackref=true,breaklinks=true,colorlinks,bookmarks=false]{hyperref}

\def\cvprPaperID{6101} % *** Enter the CVPR Paper ID here
\def\confYear{CVPR 2021}

\begin{document}

%%%%%%%%% TITLE
\title{Deep Dual Consecutive Network for Human Pose Estimation}

\author{Zhenguang Liu\\
Zhejiang Gongshang University\\
{\tt\small liuzhenguang2008@gmail.com}
% For a paper whose authors are all at the same institution,
% omit the following lines up until the closing ``}''.
% Additional authors and addresses can be added with ``\and'',
% just like the second author.
% To save space, use either the email address or home page, not both
\and
Haoming Chen\\
Zhejiang Gongshang University\\
{\tt\small chenhaomingbob@gmail.com}
\and
Runyang Feng\thanks{Corresponding Authors}\\
Zhejiang Gongshang University\\
{\tt\small huaqinew2019@gmail.com}
\and
Shuang Wu\\
Nanyang Technological University\\
{\tt\small wushuang@outlook.sg}
\and
Shouling Ji$^*$ \\
Zhejiang University\\
{\tt\small  sji@zju.edu.cn}
\and
Bailin Yang$^*$\\
Zhejiang Gongshang University\\
{\tt\small ybl@mail.zjgsu.edu.cn}
\and
Xun Wang$^*$\\
Zhejiang Gongshang University\\
{\tt\small xwang@zjgsu.edu.cn}
}

\maketitle
%\thispagestyle{empty}

%%%%%%%%% ABSTRACT
\begin{abstract}
Multi-frame human pose estimation in complicated situations is challenging. Although state-of-the-art human joints detectors have demonstrated remarkable results for static images, their performances come short when we apply these models to video sequences. Prevalent shortcomings include the failure to handle motion blur, video defocus, or pose occlusions, arising from the inability in capturing the temporal dependency among video frames. On the other hand, directly employing conventional recurrent neural networks incurs empirical difficulties in modeling spatial contexts, especially for dealing with pose occlusions. In this paper, we propose a novel multi-frame human pose estimation framework, leveraging abundant temporal cues between video frames to facilitate keypoint detection. Three modular components are designed in our framework. A Pose Temporal Merger encodes keypoint spatiotemporal context to generate effective searching scopes while a Pose Residual Fusion module computes weighted pose residuals in dual directions. These are then processed via our Pose Correction Network for efficient refining of pose estimations. Our method ranks No.1 in the Multi-frame Person Pose Estimation Challenge on the large-scale benchmark datasets \textbf{PoseTrack2017} and \textbf{PoseTrack2018}. We have released our code, hoping to inspire future research.

%Extensive experiments show that our approach significantly outperforms state-of-the-art methods on \textbf{PoseTrack2017} and \textbf{PoseTrack2018} datasets.
% achieves state-of-the-art results on both PoseTrack2017 and PoseTrack2018 benchmark datasets.
\end{abstract}

%%%%%%%%% BODY TEXT
\section{Introduction}
Human pose estimation is a fundamental problem in computer vision, which aims at locating anatomical keypoints (\emph{e.g.}, wrist, ankle, etc.) or body parts. It has enormous applications in diverse domains such as security, violence detection, crowd riot scene identification, human behavior understanding, and action recognition \cite{liu2016hierarchical}. Earlier methods \cite{wang2008multiple,wang2013beyond,zhang2009efficient,sapp2010cascaded} adopt the probabilistic graphical model or the pictorial structure model.
Recent methods have built upon the success of deep convolutional neural networks (CNNs) \cite{Cao_2017_CVPR, Toshev_2014_CVPR, Wei_2016_CVPR, Fang_2017_ICCV, newell2016stacked, chu2017multi, luo2018lstm, liu2021aggregated, liu2019towards}, achieving outstanding performance in this task. Unfortunately, most of the recent state-of-the-art methods are designed for static images, with greatly diminished performance when handling video input.

\begin{figure}[t]
\begin{center}
% \fbox{\rule{0pt}{2in} \rule{0.9\linewidth}{0pt}}
\includegraphics[width=0.95\linewidth]{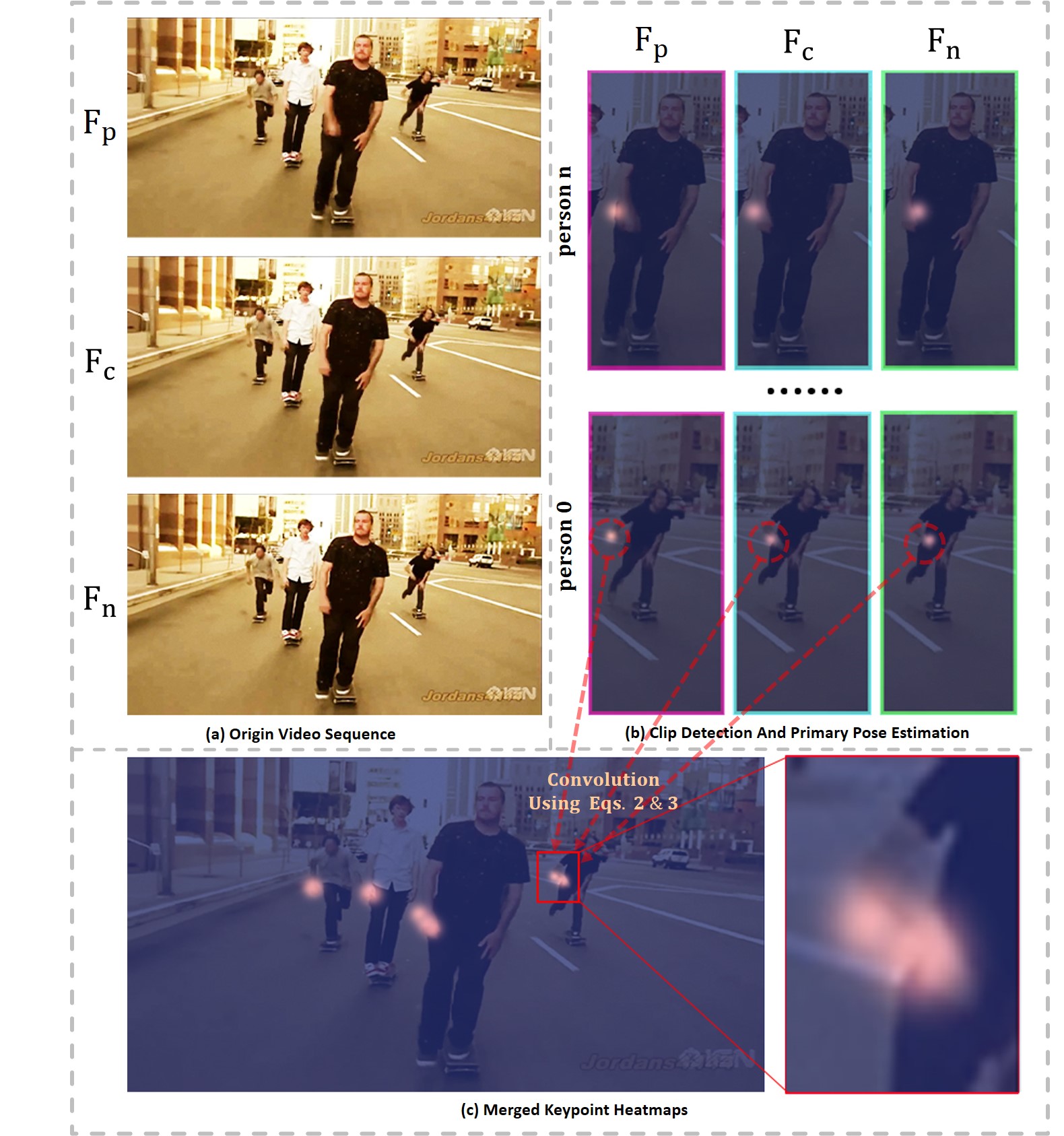}
\end{center}
\caption{An illustration of our Pose Temporal Merger (PTM) network. \textbf{(a):} Original video sequence in the datasets, and we aim to detect poses in the current frame ${F}_c$. 
\textbf{(b):} Each person in the original video sequence is assembled to a cropped clip and a single-person joints detector gives preliminary estimations of keypoint heatmaps (illustrated for right wrist).
%The rose, azure and neon borders correspond to a person in $\text{Frame}_p$ (previous), $\text{Frame}_c$ and $\text{Frame}_n$ (next) respectively.
\textbf{(c)-left:} Merged keypoint heatmaps for the right wrist, generated by our PTM network through encoding the keypoint spatial contexts. The color intensity encodes spatial aggregation.
\textbf{(c)-right:} Zoomed-in view of the merged keypoint heatmaps.}
% \label{fig:long}
\label{fig:heatmap_fuse}
\vspace{-0.5em}
\end{figure}

In this paper, we focus on the problem of multi-person pose estimation in video sequences. Conventional image-based approaches disregard the temporal dependency and geometric consistency across video frames. Dissevering these additional cues results in failure cases when dealing with challenging situations that inherently occurs in video sequences such as motion blur, video defocus, or pose occlusions. Effectively leveraging the temporal information in video sequences is of great significance to facilitate pose estimation and often plays an indispensable role for detecting heavily occluded or blurry joints.

A direct and intuitive approach to tackle this issue is to employ recurrent neural networks (RNNs) such as Long-Short Term Memory (LSTM), Gate Recurrent Unit (GRU) or 3DCNNs to model geometric consistency as well as temporal dependency across video frames. \cite{luo2018lstm} uses convolutional LSTM to capture temporal and spatial cues, and directly predicts the keypoint heatmap sequences for videos. This RNN based approach is more effective when the human subjects are spatially sparse such as single-person scenes with minimal occlusion. However, performance is severely hindered in the case of occlusion commonly occurring in multi-person pose estimation and even self-occlusion in the single-person case. \cite{wang2020combining} proposes a 3DHRNet (extension of HRNet \cite{sun2019deep} to include a temporal dimension) for extracting spatial and temporal features across video frames to estimate pose sequences. This model has shown excellent results particularly for adequately long duration single-person sequences. Another line of work considers fine-tuning the primary prediction with high confidence keypoints from the adjacent frames. \cite{song2017thin, pfister2015flowing} propose to compute the dense optical flow between every two frames, and leverage the additional flow based representations to align the predictions. This approach is promising when the optical flow can be computed precisely. However in cases involving motion blur or defocus, the poor image qualities lead to imprecise optical flows which translates to performance drops.
%This model achieves good results when working with adequately long duration single-person sequences. Again, the difficulty of obtaining a long duration single-person sequence poses is overly restrictive and poses a significant handicap for this method.

To address the shortcomings of existing methods, we propose to incorporate consecutive frames from dual temporal directions to improve pose estimation in videos. Our framework, termed Dual Consecutive network for pose estimation (DCPose), first encodes the spatial-temporal keypoint context into localized search scopes, computes pose residuals, and subsequently refines the keypoint heatmap estimations. Specifically, we design three task-specific modules within the DCPose pipeline. 1) As illustrated in Fig.~\ref{fig:heatmap_fuse}, a Pose Temporal Merger (PTM) network performs keypoints aggregation over a continuous video segment (\emph{e.g.}, three consecutive frames) with group convolution, thereby localizing the search range for the keypoint. 2) A Pose Residual Fusion (PRF) network is introduced to efficiently obtain the pose residuals between the current frame and adjacent frames. PRF computes inter-frame keypoint offsets by explicitly utilizing the temporal distance. 3) Finally, a Pose Correction Network (PCN) comprising five parallel convolution layers with different dilation rates is proposed for resampling keypoint heatmaps in the localized search range.

It is worth mentioning that the architecture of our network extends the successful PoseWarper architecture \cite{bertasius2019learning} in three ways.  (1) PoseWarper  focuses on enabling effective label propagation between frames, while we aim to refine the pose estimation of current frame using the motion context and temporal information from unlabeled neighboring frames. (2)  Information from two directions are utilized and we explicitly consider weighted residuals between frames. (3) Instead of applying the learned warping operation to a heatmap from one adjacent frame, the new network fuses together heatmaps from the adjacent frames and the current frame.

To summarize, our key contributions are: 1) A novel dual consecutive pose estimation framework is proposed. DCPose effectively incorporates bidirectional temporal cues across frames to facilitate the multi-person pose estimation task in videos. 2) We design 3 modular networks within DCPose to effectively utilise the temporal context: i) a novel Pose Temporal Merger network for effectively aggregating keypoint across frames and identifying a search scope, ii) a Pose Residual Fusion network to efficiently compute weighted pose residuals across frames, and iii) a Pose Correction Network that updates the pose estimation with the refined search scope and pose residual information. 3) Our method achieves state-of-the-art results on PoseTrack2017 and PoseTrack2018 Multi-frame Person Pose Estimation Challenge. To facilitate future research, our source code is released at \url{https://github.com/Pose-Group/DCPose}.

\begin{figure*}
\begin{center}
% \fbox{\rule{0pt}{2in} \rule{.9\linewidth}{0pt}}
\includegraphics[width=0.95\linewidth]{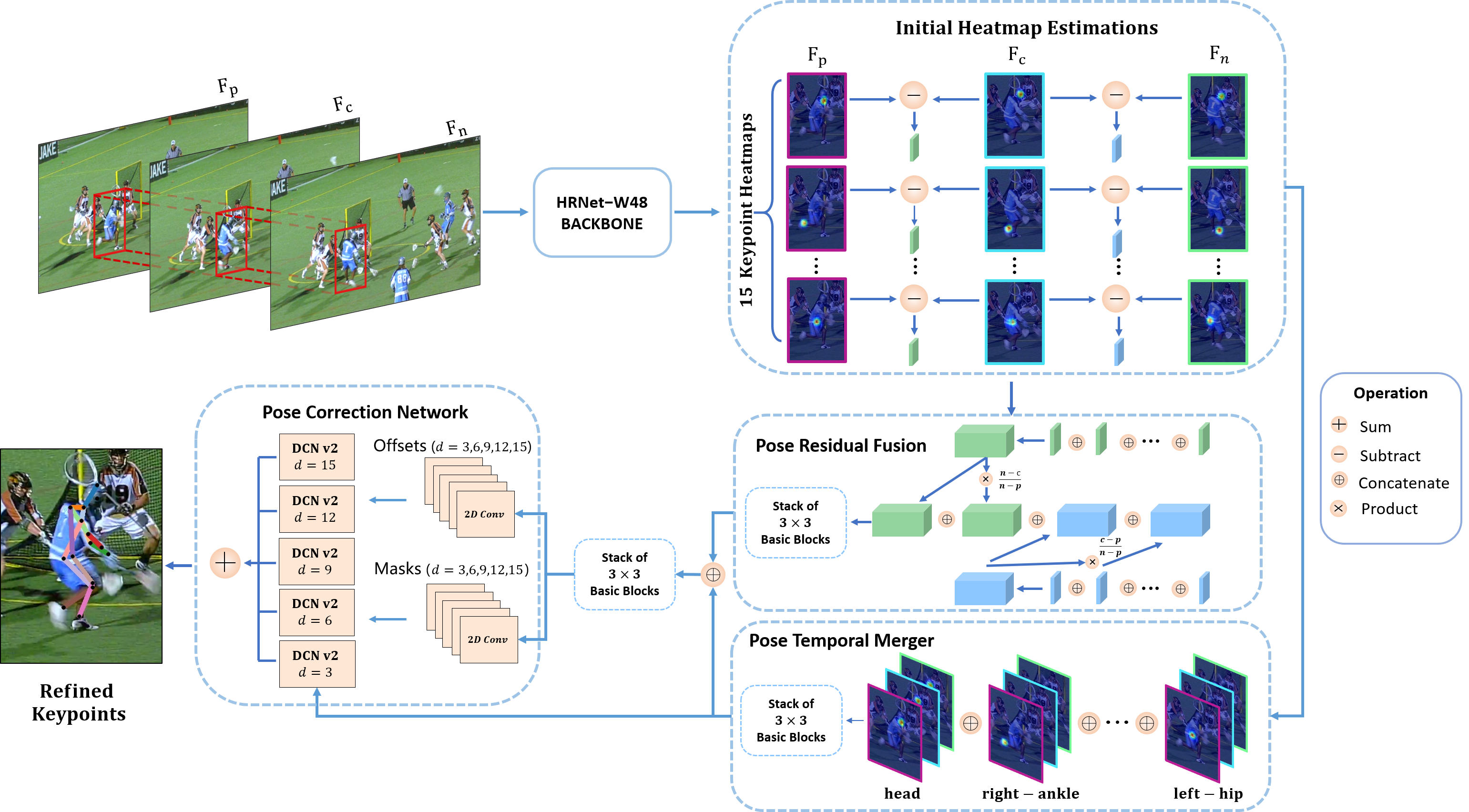}
\end{center}
\caption{Overall pipeline of our DCPose framework. The goal is to locate the keypoint positions for the current frame ${F}_c$. First, an individual person $i$ is assembled into an input sequence $\mathbf{Clip}_i(p,c,n)$, and a HRNet backbone predicts initial keypoint heatmaps $\mathbf{h}_i(p),\mathbf{h}_i(c),\mathbf{h}_i(n)$.
Our Pose Temporal Merger (PTM) and Pose Residual Fusion (PRF) networks work concurrently to obtain an effective search scope $\Phi_i(p,c,n)$ and pose residuals $\Psi_i(p,c,n)$, respectively. These are then fed into our Pose Correction Network (PCN) which refines the keypoint estimation for person $i$ in ${F}_c$.}
%PCN consists of five $3\times3$ dilated convolution layers to calculate the offsets $O(p_i)$ for all pixel locations $p_i$ and masks $M(s_i)$ for all sampling point locations $s_i$ after offsetting, the atrous rates $d\in\{3,6,9,12,15\}$. The offsets and masks are used to resample the final keypoint heatmaps. All five predictions are aggregated to generate the pose heatmaps of $\text{Frame}_c$.}
\label{fig:pipeline}
\end{figure*}
%------------------------------------------------------------------------- 

\section{Related Work}
\subsection{Imaged Based Multi Person Pose Estimation}
Earlier image-based human pose estimation works generally fall within a pictorial structure model paradigm, in which the human body is represented as a tree-structured model \cite{wang2008multiple,wang2013beyond,zhang2009efficient,sapp2010cascaded} or a forest model \cite{sun2012conditional,dantone2013human}. Despite allowing for efficient inference, these approaches tend to be insufficient in modeling complex relationships between body parts, and this weakness is accentuated when temporal information enters the picture. %These approaches have been demonstrated to be very efficient in human pose estimation.
Recently, neural networks based methods \cite{lin2018learning, liu2015graph, tang2019learning, artacho2020unipose, cheng2020higherhrnet, huang2020devil, zhang2020distribution, varamesh2020mixture, su2019multi, Chengzhiyong, ijcai2020} have been in the spotlight due to their superior performance in various fields. One line of work \cite{carreira2016human} outputs skeletal joints coordinates directly by regressing image features. Another approach \cite{newell2016stacked, pfister2015flowing} utilises probability heatmaps to represent joints locations. Due to the reduced difficulty for optimization, heatmap based pose estimation have since been widely adopted. In general, these methods can be classified into part-based framework (bottom-up) and two-step framework (top-down). The bottom-up approach \cite{Cao_2017_CVPR, kocabas2018multiposenet, kreiss2019pifpaf,li2019crowdpose} first detects individual body parts, then assembles these constituent parts into the entire person. \cite{Cao_2017_CVPR} builds a bottom-up pipeline and utilizes part affinity fields to capture pairwise relationships between different body parts. Conversely, the top-down approach \cite{Fang_2017_ICCV, xiao2018simple, Wei_2016_CVPR, sun2019deep, newell2016stacked, moon2019posefix} first performs person detection, then proceeds with single-person pose estimation on each individual. \cite{Wei_2016_CVPR} proposes a sequential architecture of convolutional pose machines, which follows a strategy of iteratively refining the output of each network stage. \cite{Fang_2017_ICCV} designs a symmetric spatial transformer network for extracting a high-quality single person region from an inaccurate bounding box. A recent work in \cite{sun2019deep} proposes a HRNet that performs multi-scale fusion to retain high resolution feature maps. This improves spatial precision in keypoint heatmaps and achieves the state-of-the-art on several image-based benchmarks.

\subsection{Video Based Multi Person Pose Estimation}
Directly applying the existing image-level methods to video sequences produces unsatisfactory predictions, primarily due to the failure to capture temporal dependency among video frames. Consequently, these models fail to handle motion blur, video defocus, or pose occlusions which are frequently encountered in video inputs. \cite{pfister2015flowing, song2017thin, weinzaepfel2013deepflow} compute the dense optical flow between every consecutive frames with the flow representations providing additional cues for aligning predictions. However, motion blur, defocus or occlusion occurrences hinder optical flow computation and affect performance. \cite{luo2018lstm} replaces the convolutional pose machines in \cite{Wei_2016_CVPR} with convolutional LSTMs for modeling temporal information in addition to spatial contexts. A principal shortcoming of such an approach is being severely impacted by occlusion. \cite{bertasius2019learning} proposes to learn an effective video pose detector from sparsely labeled videos through a warping mechanism and has turned out to be very successful,  dominating the PoseTrack leaderboard for a long time. \cite{wang2020combining} extends HRNet \cite{sun2019deep} with temporal convolutions and proposes 3DHRNet, which is successful in handling pose estimation and tracking jointly. %, which presents a major drawback.
%3DHRNet, relies however on sufficiently long single person sequences for favorable performance. %, which presents a major drawback.

\section{Our Approach}\label{section:method}
The pipeline of our proposed DCPose is illustrated in Fig.~\ref{fig:pipeline}. To improve keypoint detection for the current frame ${F}_c$, we make use of additional temporal information from a previous frame ${F}_p$ and a future frame ${F}_n$. ${F}_p$ and ${F}_n$ are selected within a frame window $[c-T,c+T]$, where $p \in [c-T, c)$ and $n \in (c, c+T]$ respectively denote the frame indices. The bounding boxes for individual persons in $F_c$ are first obtained by a human detector. Each bounding box is enlarged by 25\% and is further used to crop the same person in $F_p$ and ${F}_n$. Individual $i$ in the video will thus be composed of a cropped video segment, which we denote as $\mathbf{Clip}_i(p,c,n)$.
$\mathbf{Clip}_i(p,c,n)$ is then fed into a backbone network that serves to output preliminary keypoint heatmap estimations $\mathbf{h}_i(p,c,n)$. The pose heatmaps $\mathbf{h}_i(p,c,n)$ is then processed in parallel through two modular networks, Pose Temporal Merger (PTM) and Pose Residual Fusion (PRF). PTM outputs $\Phi_i(p,c,n)$, {which encodes the spatial aggregation}, and PRF computes $\Psi_i(p,c,n)$, {which captures pose residuals in two directions}. Both feature tensors $\Phi_i(p,c,n)$ and $\Psi_i(p,c,n)$ are then simultaneously fed into our Pose Correction Network (PCN) to refine and improve upon the initial pose estimations. In what follows, we introduce the three key components in detail.

\subsection{Pose Temporal Merger}
The motivation for our Pose Temporal Merger (PTM) comes from the following observations and heuristics. 1) Although existing pose estimation methods such as \cite{sun2019deep, Fang_2017_ICCV} suffer from performance deterioration on videos, we observe that their predictions do still provide useful information for approximating the keypoint spatial positions. 2) Another heuristic is on temporal consistency, \emph{i.e.,} the pose of an individual does not undergo dramatic and abrupt changes across a very few frame intervals (typically 1/60 to 1/25 of a second per frame). Therefore, we design PTM to encode the keypoint spatial contexts based on initial predictions (from a backbone network), providing a compressed search scope that facilitates refinement and correction of pose prediction within a confined range.

For person $i$, the backbone network returns initial keypoint heatmaps $\mathbf{h}_i(p),\mathbf{h}_i(c),\mathbf{h}_i(n)$. Naively, we could merge them through a direct summation $\mathbf{H}_i(p,c,n)=\mathbf{h}_i(p)+\mathbf{h}_i(c)+\mathbf{h}_i(n)$. However, we expect that the additional information that may be extracted from ${F}_p$ and ${F}_n$ is inversely proportional to their temporal distances from current frame ${F}_c$. We formalize this intuition as:
\begin{equation}
\begin{aligned}
\mathbf{H}(p,c,n) = \frac{n-c}{n-p}\mathbf{h}_i(p)+\mathbf{h}_i(c)+\frac{c-p}{n-p}\mathbf{h}_i(n). \label{eqn:mergedheatmap}
\end{aligned}
\end{equation}
Recall that $p,c,n$ are frame indices. We explicitly assign higher weights to the frames that are temporally nearer to the current frame.

Based on the important fact that convolution operations serve to adjust (feature) weights, we utilise convolutional neural networks to practically implement the idea  %expedite and parallelize the 
%computations for the merged keypoint heatmaps 
of Eq.~\ref{eqn:mergedheatmap}. However, including all joint channels in the computation for the merged keypoint heatmap of a single joint will result in redundancy. For example, when encoding the spatial context of the left wrist, involving other joints such as the head and ankle at different times will likely not to have any bearing and may even breed confusion. %obscure useful information. 
% and . 
Therefore, for each joint, we only include its own specific temporal information for computing its merged keypoint heatmap. This is implemented via a group convolution. We regroup the keypoint heatmaps $\mathbf{h}_i(p),\mathbf{h}_i(c),\mathbf{h}_i(n)$ according to joint, and stack them to a feature tensor $\phi_i$, which can be expressed as:
\begin{equation}
\begin{aligned}
\phi_i(p,c,n) = \bigoplus_{j=1}^{N} \frac{n-c}{n-p}\mathbf{h}_i^j(p)\oplus\mathbf{h}_i^j(c)\oplus\frac{c-p}{n-p}\mathbf{h}_i^j(n)
\end{aligned}
\end{equation}
where $\oplus$ denotes the concatenate operation and the superscript $j$ index the $j$-th joint for a total of $N$ joints. Subsequently, the feature tensor $\phi_i$ is fed into a stack of 3 $\times$ 3 residual blocks (adapted from the residual steps block in RSN \cite{cai2020learning}), producing the merged keypoint heatmaps $\Phi_i(p,c,n)$:
\begin{equation}
\begin{aligned}
\phi_i(p,c,n) \xrightarrow[\text{residual blocks}]{\text{stack of }3\times3} \Phi_i(p,c,n).
\end{aligned}
\end{equation}
This group convolution not only eliminates the disturbance of irrelevant joints, but also removes redundancy and shrinks the amount of model parameters required. It is also advantageous to directly summing keypoint heatmaps in Eq.~\ref{eqn:mergedheatmap} since the group CNN operation allow different weights %distribution 
at the pixel level and benefits learning an end-to-end model.
Visual results of aggregated keypoint heatmaps following our PTM is illustrated in Fig.~\ref{fig:heatmap_fuse}.

\subsection{Pose Residual Fusion}
\label{res}
Parallel to spatial aggregation of keypoint heatmaps in PTM, our Pose Residual Fusion (PRF) branch aims to compute the pose residuals which will serve as additional favorable temporal cues. Given keypoint heatmaps $\mathbf{h}_i(p),\mathbf{h}_i(c),\mathbf{h}_i(n)$, we compute the pose residual features as follows:
\begin{equation}
\begin{aligned}
\psi_i(p,c)&=\mathbf{h}_i(c)-\mathbf{h}_i(p)\\
\psi_i(c,n)&=\mathbf{h}_i(n)-\mathbf{h}_i(c)\\
\psi_i&=\psi_i(p,c)\oplus\psi_i(c,n)\\
&\oplus\frac{n-c}{n-p}\psi_i(p,c)\oplus\frac{c-p}{n-p}\psi_i(c,n).
\end{aligned}
\end{equation}
$\psi_i$ concatenates the original pose residuals $\psi_i(p,c)$, $\psi_i(c,n)$, and their weighted versions, where the weights are obtained according to the temporal distances. Similar to PTM, $\psi_i$ is then processed via a stack of $3\times3$ residual blocks to give the final pose residual feature $\Psi_i(p,c,n)$:
\begin{equation}
\begin{aligned}
\psi_i(p,c,n) \xrightarrow[\text{residual blocks}]{\text{stack of }3\times3} \Psi_i(p,c,n).
\end{aligned}
\end{equation}

\subsection{Pose Correction Network}
Given the merged keypoint heatmaps $\Phi_i(p,c,n)$ and pose residual feature tensor $\Psi_i(p,c,n)$, our Pose Correction Network is employed to refine the initial keypoint heatmap estimation $\mathbf{h}_i(c)$, yielding adjusted final keypoint heatmaps. Primarily, the pose residual feature tensor $\Psi_i(p,c,n)$ is used as the input for five parallel $3\times3$ convolution layers with different dilation rates $d\in\{3,6,9,12,15\}$. This computation gives five groups of offsets for the five kernels of the subsequent deformable convolution layer. Formally, the offsets are computed as:
%pose offsets $O_d$ which is defined for all pixel locations
\begin{equation}
\begin{aligned}
\Phi_i(p,c,n)\oplus\Psi_i(p,c,n) \xrightarrow[\text{residual blocks}]{\text{stack of }3\times3}\xrightarrow[\text{convolution layers}]{\text{dilation rate } d} O_{i,d}.
\end{aligned}
\end{equation}
Different dilation rates correspond to varying the size of the effective receptive field whereby enlarging the dilation rate \cite{yu2015multi} increases the scope of the receptive field. A smaller dilation rate focuses on local appearance, which is more sensitive for capturing subtle motion contexts.
Conversely, using a large dilation rate allows us to encode global representations and capture relevant information of a larger spatial scope. In addition to the offset computation, we feed the merged keypoint heatmaps to similar convolution layers and obtain five sets of masks $M_d$ as:
\begin{equation}
\begin{aligned}
\Phi_i(p,c,n)\oplus\Psi_i(p,c,n) \xrightarrow[\text{residual blocks}]{\text{stack of }3\times3}\xrightarrow[\text{convolution layers}]{\text{dilation rate } d} M_{i,d}.
% \Phi_i(p,c,n)\oplus\Psi_i(p,c,n) \xrightarrow[\text{residual blocks}]{\text{stack of }3\times3} \xrightarrow[\text{convolution layers}]{\text{dilation rate } d} M_{i,d}.
\end{aligned}
\end{equation}
%The support of $M_d$ is the set of all sampling point locations after offsetting. 
The parameters of two dilation convolution structures for offset $O$ and mask $M$ computation are independent. A mask $M_d$ can be considered as the weight matrix for a convolution kernel.

We implement the pose correction module through the deformable convolution $V2$ network (DCN v2 \cite{zhu2019deformable}) at various dilation rates $d$. DCN v2 takes the following inputs:
1) the merged keypoint heatmaps $\Phi_i(p,c,n)$, 2) the kernel offsets $O_{i,d}$, and 3) the masks $M_{i,d}$, and outputs a pose heatmap for person $i$ at dilation rate $d$:
\begin{equation}
\begin{aligned}
\left(\Phi_i(p,c,n),O_{i,d},M_{i,d}\right)\xrightarrow[\text{DCN v2}]{\text{dilation rate } d} \mathbf{H}_{i,d}(c).
\end{aligned}
\end{equation}
The five outputs for five dilation rates are summarized and normalized to yield the final pose prediction for person $i$:
\begin{equation}
\begin{aligned}
\sum_{d\in\{3,6,9,12,15\}}\mathbf{H}_{i,d}(c)\xrightarrow[]{\text{normalization}} \mathbf{H}_{i}(c).
\end{aligned}
\end{equation}
Ultimately, the above procedure is performed for each individual $i$. By effectively utilising the additional cues from $F_p$ and ${F}_n$ in our DCPose framework, the final pose heatmaps are enhanced and improved.

\subsection{Implementation Details}
\textbf{Backbone Model}\quad
Our network is highly adaptable and we can seamlessly integrate any image based pose estimation architecture as our backbone. We employ the state-of-the-art Deep High Resolution Network (HRNet-W48 \cite{sun2019deep}) as our backbone joints detector, since its superior performance for single image pose estimation will be beneficial for our approach.
\renewcommand\arraystretch{1.2}
\begin{table}
  \resizebox{0.48\textwidth}{!}{
  \begin{tabular}{l|c|c|c|c|c|c|c|c}
    % \hline
    \hline
      Method                            &Head   &Shoulder &Elbow       &Wrist   &Hip    &Knee   &Ankle   &{\bf Mean}\cr
    \hline
     PoseFlow\cite{xiu2018pose}         &$66.7$ & $73.3$  &$68.3$      &$61.1$  &$67.5$ &$67.0$ &$61.3$  &{$\bf 66.5$}\cr
JointFlow\cite{doering2018joint}        & -     & -       &-           &-       &-      &-      &-       &{\bf $\bf 69.3$}\cr
     FastPose\cite{zhang2019fastpose}   &$80.0$ &$80.3$   &$69.5$      &$59.1$  &$71.4$ &$67.5$ &$59.4$  &{$\bf 70.3$}\cr
SimpleBaseline\cite{xiao2018simple}     &$81.7$ &$83.4$   &$80.0$      &$72.4$  &$75.3$ &$74.8$ &$67.1$  &{$\bf 76.7$}\cr
  STEmbedding\cite{jin2019multi}        &$83.8$ &$81.6$   &$77.1$      &$70.0$  &$77.4$ &$74.5$ &$70.8$  &{$\bf 77.0$}\cr
        HRNet\cite{sun2019deep}         &$82.1$ &$83.6$   &$80.4$      &$73.3$  &$75.5$ &$75.3$ &$68.5$  &{$\bf 77.3$}\cr
         MDPN\cite{guo2018multi}        &$85.2$ &$88.5$   &$83.9$      &$77.5$  & $79.0$&$77.0$ &$71.4$  &{$\bf 80.7$}\cr
 PoseWarper\cite{bertasius2019learning} &$81.4$ &$88.3$   &$83.9$      &$ 78.0$ &$82.4$ &$80.5$ &$73.6$  &{ $\bf 81.2$}\cr
    \hline
   \bf DCPose  &$\bf 88.0$  &$\bf 88.7$     &$\bf 84.1$   &$\bf 78.4$&$\bf 83.0$        &$\bf 81.4$&$\bf 74.2$      &$\bf 82.8$\cr
    \hline
    \end{tabular}}
    \vspace{0.2em}\caption{\footnotesize{Quantitative Results (\textbf{AP}) on  PoseTrack2017 \bf validation set}.}  \label{tab1}
\end{table}

\renewcommand\arraystretch{1.2}
\begin{table}
  \resizebox{0.48\textwidth}{!}{
  \begin{tabular}{l|c|c|c|c|c|c|c|c}
    \hline
     Method                            &Head&Shoulder &Elbow  &Wrist &Hip &Knee &Ankle &{\bf Total}\cr
    \hline
PoseFlow\cite{xiu2018pose}             &$64.9$  &$67.5$&$65.0$ &$59.0$ &$62.5$ &$62.8$  &$57.9$   &{\bf $\bf 63.0$}\cr
JointFlow\cite{doering2018joint}       &-       &-     &-      &$53.1$ &-      &-       &$50.4$   &{$\bf 63.4$}\cr
KeyTrack\cite{snower202015}            &-       &-     &-      &$71.9$ &-      &-       &$65.0$   &{$\bf 74.0$}\cr
DetTrack\cite{wang2020combining}       &-       &-     &-      &$69.8$ &-      &-       &$65.9$   &{$\bf 74.1$}\cr
SimpleBaseline\cite{xiao2018simple}    &$80.1$ &$80.2$ &$76.9$ &$71.5$ &$72.5$ &$72.4$  &$65.7$   &{$\bf 74.6$}\cr
HRNet\cite{sun2019deep}                &$80.1$ &$80.2$ &$76.9$ &$72.0$ &$73.4$ &$72.5$  &$67.0$   &{$\bf 74.9$}\cr
PoseWarper\cite{bertasius2019learning} &$79.5$ &$84.3$ &$80.1$ &$75.8$ &$77.6$ &$76.8$  &$70.8$   &$\bf 77.9$\cr
    \hline
    \bf DCPose&$\bf 84.3$&$\bf 84.9$&$\bf 80.5$&$\bf 76.1$&$\bf 77.9$&$\bf 77.1$&$\bf 71.2$&$\bf 79.2$\cr
    \hline
    \end{tabular}}
    \vspace{0.2em}\caption{\footnotesize{Performance comparisons on the PoseTrack2017 \textbf{test} set.}} \label{tab2}
\end{table}

\renewcommand\arraystretch{1.2}
\begin{table}
   \resizebox{0.48\textwidth}{!}{
   \begin{tabular}{l|c|c|c|c|c|c|c|c}
     \hline
      Method                            &Head &Shoulder &Elbow  &Wrist &Hip &Knee &Ankle &{\bf Mean}\cr
     \hline
 AlphaPose\cite{Fang_2017_ICCV}         &$63.9$  &$78.7$&$77.4$ &$71.0$ &$73.7$ &$73.0$    &69.7     &{\bf $\bf 71.9$}\cr
 MDPN\cite{guo2018multi}                &$75.4$ &$81.2$ &$79.0$ &$74.1$ &$72.4$ &$73.0$  &$69.9$   &{$\bf 75.0$}\cr
         PoseWarper\cite{bertasius2019learning} &$79.9$&$86.3$&$82.4$&$77.5$&$79.8$&$78.8$&$73.2$&{ $\bf 79.7$}\cr
     \hline
     \bf DCPose&$\bf 84.0$&$\bf 86.6$&$\bf 82.7$&$\bf 78.0$&$\bf 80.4$&$\bf 79.3$&$\bf 73.8$&$\bf 80.9$\cr
     \hline
     \end{tabular}}
     \vspace{0.2em}\caption{\footnotesize{Quantitative Results(\textbf{AP}) on  PoseTrack2018 \bf validation set}.} \label{tab3}
   \end{table}

\renewcommand\arraystretch{1.2}
\begin{table}
   \resizebox{0.48\textwidth}{!}{
   \begin{tabular}{l|c|c|c|c|c|c|c|c}
     \hline
      Method                            &Head &Shoulder &Elbow  &Wrist &Hip &Knee &Ankle &{\bf Total}\cr
     \hline
 AlphaPose++\cite{Fang_2017_ICCV,guo2018multi}         &- &- &- &$66.2$ &- &-  &$65.0$     &{\bf $\bf 67.6$}\cr
 DetTrack \cite{wang2020combining}      &-      &-      &-      &$69.8$ &-      &-       &$67.1$     &{\bf $\bf 73.5$}\cr
 MDPN\cite{guo2018multi}                &- &- &- &74.5 &- &-  &69.0   &{$\bf 76.4$}\cr
PoseWarper\cite{bertasius2019learning} &$78.9$&$\bf 84.4$&$\bf 80.9$&$76.8$&$75.6$&$77.5$&$71.8$&{ $\bf 78.0$}\cr
     \hline
     \bf DCPose&$\bf 82.8$&$ 84.0$&$ 80.8$&$\bf 77.2$&$\bf 76.1$&$\bf 77.6$&$\bf 72.3$&$\bf 79.0$\cr
     \hline
     \end{tabular}}
     \vspace{0.2em}\caption{\footnotesize{Performance comparisons on the PoseTrack2018 \textbf{test} set.}} \label{tab4}
   \end{table}
\textbf{Training}\quad
Our Deep Dual Consecutive Network is implemented in PyTorch. During training, we use the ground truth person bounding boxes to generate the $\mathbf{Clip}_i(p,c,n)$ for person $i$ as the input sequence to our model. For boundary cases, we apply same padding. In other words, if there are no frames to extend forward and backward from ${F}_c$,
${F}_p$ or ${F}_n$ will be replaced by ${F}_c$. We utilize the HRNet-W48 pretrained on the PoseTrack dataset as our backbone, and freeze the backbone parameters throughout training, only backpropagating through the subsequent components in DCPose.
%Considering the group convolution that we assume the parameter $groups$ of the convolution layer should be equal to the joints amounts, the reason why we attach 17 to the parameter $groups$ is that the keypoints sequence of the MSCOCO dataset has 17 joints. To resample the final pose heatmap, we employ five $3\times3$ deformable convolution layers, which also utilize five different dilation rates d$\in\{3,6,12,18,24\}$. Each deformable convolution layer is used to predict a pose heat maps result, the five results are medially aggregated to obtain the final pose heat maps.

\textbf{Loss function}\quad
We employ the standard pose estimation loss function as our cost function. Training aims to minimize the total Euclidean or $L2$ distance between prediction and ground truth heatmaps for all joints. The cost function is defined as:
\begin{equation}
\begin{aligned}
L = \frac{1}{N} * \sum_{j=1}^N v_j * ||G(j) - P(j)||^2
\end{aligned}
\end{equation}
Where $G(j)$, $P(j)$ and $v_j$ respectively denote the ground truth heatmap, prediction heatmap and visibility for joint $j$. During training, the total number of joints is set to $N=15$. The ground truth heatmaps are generated via a 2D Gaussian centered at the joint location.
\section{Experiments}
In this section, we present our experimental results on two large-scale benchmark datasets: Posetrack2017 and PoseTrack2018 Multi-frame Person Pose Estimation Challenge datasets. %Notably, our approach achieves state-of-the-art results for both two benchmark datasets.

\subsection{Experimental Settings}
\textbf{Datasets}\quad
PoseTrack is a large-scale public dataset for human pose estimation and articulated tracking in video and includes challenging situations with complicated movement of highly occluded people in crowded environments. The PoseTrack2017 dataset \cite{Iqbal_2017_CVPR} contains 514 video clips and 16,219 pose annotations, with 250 clips for training, 50 clips for validation, and 214 clips for testing. The PoseTrack2018 dataset greatly increased the number of video clips to a total of 1,138 clips and 153,615 pose annotations. The training, validation, and testing splits consist of 593, 170, and 375 clips respectively. The 30 frames in the center of the training video clips are densely annotated. For the validation clips, annotations are provided every four frames. Both PoseTrack2017 and PoseTrack2018 identify 15 joints, with an additional annotation label for joint visibility. We evaluate our model only for visible joints with the average precision (\textbf{AP}) metric \cite{Iqbal_2017_CVPR, Andriluka_2018_CVPR}.
\renewcommand\arraystretch{1.1}
\begin{table*}
\centering
\resizebox{1\textwidth}{!}{
\begin{tabular}{l|c|c|c|c|c|c|c|c}
\hline
Method & Head & Shoulder & Elbow & Wrist & Hip & Knee & Ankle & {\bf Mean}\cr
\hline
\textbf{DCPose, complete, $T=1$}
& $\bf 88.0$ & $\bf 88.7$ & $\bf 84.1$ & $\bf 78.4$ & $\bf 83.0$ & $\bf 81.4$ & $\bf 74.2$ & $\bf 82.8$ \cr
\hline
r/m PTM, $\Phi(p,c,n)\leftarrow\mathbf{h}(p)+\mathbf{h}(c)+\mathbf{h}(n)$
& 87.3 & 88.4 & 83.6 & 77.6 & 82.8 & 78.4 & 73.7 & 82.0 \cr
r/m PTM, $\Phi(p,c,n)\leftarrow\mathbf{h}(c)$
& 87.5 & 88.5 & 83.7 & 78.1 & 82.9 & 80.8 & 73.4 & 82.2 \cr
\hline
r/m PRF
& 87.4 & 88.3 & 83.6 & 77.6 & 82.7 & 80.5 & 73.4 & 82.1 \cr
\hline     
%  no dilated convolutions, $d=1$                               &-          &-           &-          &$53.1$   &-              &-             &$50.4$      &{$\bf 63.4$}\cr
1 dilation convolution, $d=3$
& 87.3 & 88.5 & 83.8 & 72.6 & 82.9 & 78.4 & 73.7 & 81.7 \cr
2 dilation convolutions, $d\in\{3,6\}$
& 87.5 & 88.2 & 83.6 & 74.7 & 82.8 & 79.8 & 73.6 & 81.9 \cr
3 dilation convolutions, $d\in\{3,6,9\}$
& 87.9 & 88.6 & 83.9 & 78.3 & 82.9 & 81.4 & 74.0 & 82.7 \cr
4 dilation convolutions , $d\in\{3,6,9,12\}$
& 87.9 & 88.6 & 84.0 & 76.9 & 83.0 & 80.9 & 74.0 & 82.6 \cr
%5 dilation convolutions, $d\in\{3,6,9,12,15\}$ & 88.4 & 88.8 & 84.1 & 78.4 & 83.0 & 81.5 & 74.3 & 83.0 \cr
\hline
r/m previous frame
& 87.8 & 88.4 & 83.8 & 77.8 & 82.6 & 80.6 & 73.7 & 82.2 \cr
r/m next frame
& 88.0 & 88.6 & 83.7 & 77.7 & 82.8 & 80.7 & 73.2 & 82.3 \cr
% r/m group convolutions                                        &-       &-     &-      &$53.1$ &-      &-       &$50.4$   &{$\bf 63.4$}\cr
% \hline
% r/m temporal distance                                         &-       &-     &-      &$53.1$ &-      &-       &$50.4$   &{$\bf 63.4$}\cr
\hline
$T=2$
& 87.7 & 88.4 & 83.8 & 78.0 & 83.0 & 80.8 & 73.2 & 82.6 \cr
$T=3$
& 87.6 & 88.4 & 83.6 & 77.7 & 82.6 & 80.7 & 71.7 & 82.1 \cr
$T=4$
& 81.2 & 87.4 & 83.3 & 76.7 & 81.9 & 80.2 & 72.1 & 81.7 \cr
\hline
\end{tabular}}   
\vspace{0.2em}\caption{\normalsize{Ablation study of different components in our DCPose performed on PoseTrack2017 validation set. ``r/m X" refers to removing X module in our network. The complete DCPose consistently achieves the best results which are highlighted.}} \label{tab5}
\end{table*} 

\textbf{Parameter Settings}\quad
During training, we incorporate data augmentation including random rotations, scaling, truncating, and horizontal flipping to increase variation. Input image size is fixed to $384\times288$. The default interval between ${F}_c$ and ${F}_p$ or ${F}_n$ is set to 1. Backbone parameters are fixed to the pretrained HRNet-W48 model weights.
All subsequent weight parameters are initialized from a Gaussian distribution with $\mu=0$ and $\sigma=0.001$, while bias parameters are initialized to $0$. We employ the Adam optimizer with a base learning rate of $0.0001$ that decays by $10\%$ every 4 epochs. We train our model for a batch size of 32 for 20 epochs with 2 Nvidia GeForce Titan X GPUs.

\subsection{Comparison with State-of-the-art Approaches}
\textbf{Results on the PoseTrack2017 Dataset}\quad
We evaluate our approach on PoseTrack2017 validation set and full test set using the widely adopted average precision (\textbf{AP}) metric  \cite{xiao2018simple,xiu2018pose,doering2018joint,zhang2019fastpose}.
% Firstly the accuracy of each joint is calculated, and then we take the average of all joints precisions as the final performance of the models.
Table \ref{tab1} presents the quantitative results of different approaches in terms of AP on PoseTrack2017 validation set.
We benchmark our \textbf{DCPose} model against eight existing methods \cite{xiu2018pose}, \cite{doering2018joint}, \cite{zhang2019fastpose},  
\cite{xiao2018simple}, \cite{jin2019multi}, \cite{sun2019deep}, \cite{guo2018multi}, \cite{bertasius2019learning}. In Table \ref{tab1}, the APs of key joints, such as Head, Shoulder, Knee, and Elbow, are reported, as well as the mAP (mean AP) for all joints. 
%We benchmark our \textbf{DCPose} model against eight existing methods \textbf{PoseFlow} \cite{xiu2018pose}, \textbf{JointFlow} \cite{doering2018joint}, \textbf{FastPose} \cite{zhang2019fastpose}, 
%\textbf{SimpleBaseline} \cite{xiao2018simple}, \textbf{STEmbedding} \cite{jin2019multi}, \textbf{HRNet} \cite{sun2019deep}, \textbf{MDPN} \cite{guo2018multi}, \textbf{PoseWarper} \cite{bertasius2019learning}. In Table \ref{tab1}, the APs of key joints, such as Head, Shoulder, Knee, and Elbow, are reported, as well as the mAP for all joints. 

\begin{figure*}
\begin{center}
% \fbox{\rule{0pt}{2in} \rule{.9\linewidth}{0pt}}
\includegraphics[width=1\linewidth]{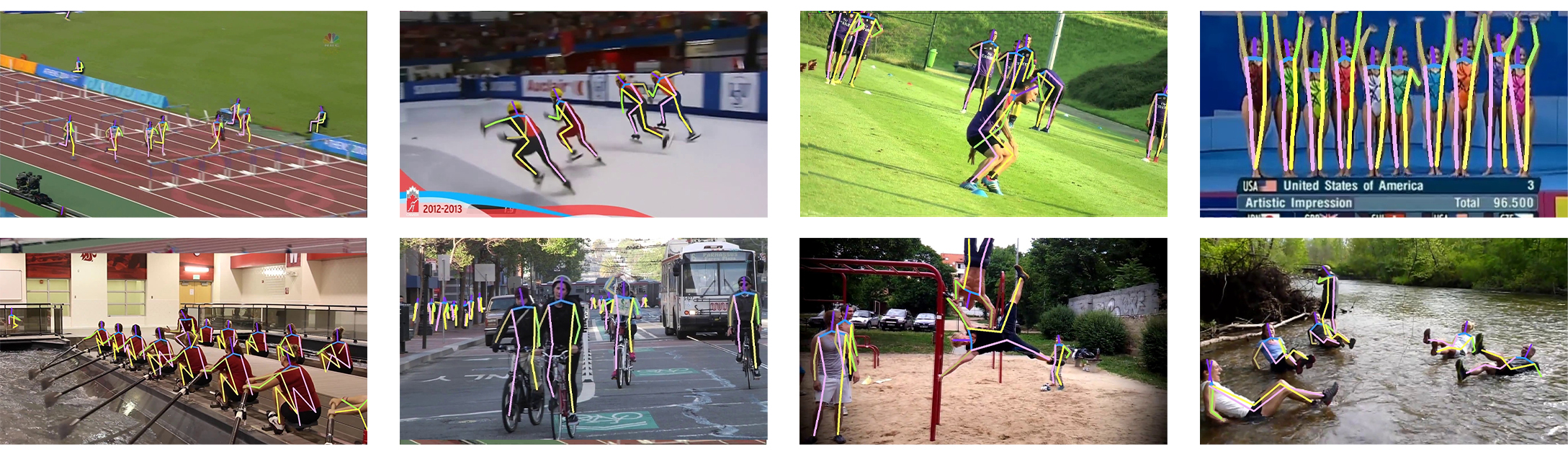}
\end{center}
\caption{Visual results of our model on the PoseTrack2017 and PoseTrack2018 datasets comprising complex scenes: high-speed movement, occlusion, and multiple persons.}
\label{fig:results}
\end{figure*}

Results on the test set are provided in Table \ref{tab2}. These results are obtained by uploading our prediction results to the  PoseTrack evaluation server:\url{https://posetrack.net/leaderboard.php} because the annotations for test set are not public. Our DCPose network achieves state-of-the-art results for multi-frame person pose estimation challenge for both the validation and test sets. DCPose consistently  outperforms existing methods and achieves a \textbf{mAP} of 79.2. The performance boost for relatively difficult joints is also encouraging: we obtain an \textbf{mAP} of 76.1 for the wrist and an \textbf{mAP} of 71.2 for the ankle. Some sample results are displayed in Fig.~\ref{fig:results}, which are indicative of the effectiveness of our method in complex scenes. More visualized results can be found in \url{https://github.com/Pose-Group/DCPose}.

{\textbf{Results on the PoseTrack2018 Dataset}}\quad
We also evaluate our model on the PoseTrack2018 dataset. The validation and test set \text{AP} results are tabulated in Table \ref{tab3} and Table \ref{tab4}, respectively. As shown in the tables, our approach once again delivers the state-of-the-art results. We achieve an \textbf{mAP} of 79.0 on the test set, and obtain an \textbf{mAP} of 77.2 for the difficult \emph{wrist} joint and 72.3 for the \emph{ankle}.
%-------------------------------------------------------------------------

\subsection{Ablation Experiments}
Extensive ablation experiments are performed on the PoseTrack2017 dataset to study the effectiveness of various components in our DCPose framework. Through ablating the various modular networks including Pose Temporal Merger (PTM), Pose Residual Fusion (PRF) and Pose Correction Network, we evaluate their contributions towards the overall performance. We also investigate the efficacy of including information from both temporal directions as well as the impact of modifying the temporal distance, \emph{i.e.,} frame intervals between ${F}_c$ and ${F}_p$, ${F}_n$. The results are reported in Table~\ref{tab5}.

\textbf{Pose Temporal Merger}\quad
For this ablation setting, we remove the PTM and instead obtain the merged pose heatmaps $\Phi(p,c,n)$ as follows:
i) $\mathbf{h}(p)+\mathbf{h}(c)+\mathbf{h}(n)\xrightarrow[\text{convolution layer}]{3\times3}\Phi(p,c,n)$; ii) $\mathbf{h}(c)\xrightarrow[]{}\Phi(p,c,n)$. The \textbf{mAP} falls from 82.8 to 82.0 for (i) and 82.2 for (ii). This significant performance degradation upon removal of the PTM module can be attributed to the failure of obtaining an effective search range for the joints, which lead to decreased accuracy in locating the joint in the subsequent pose correction stage.

\textbf{Pose Residual Fusion}\quad
We investigate removing the PRF and compute the pose residual maps $\Psi(p,c,n)$ with the following scheme:
$\mathbf{h}(c)-\mathbf{h}(p)\oplus\mathbf{h}(n)-\mathbf{h}(c)\xrightarrow[\text{convolution layer}]{3\times3}\Psi(p,c,n)$. This results in the \textbf{mAP} dropping 0.7 to 82.1. This significant reduction in performance highlights the important role of PRF in providing accurate pose residual cues for computing offsets and guiding the keypoint localization. %We present the pose residual heatmaps in Fig.~\ref{fig:pipeline}.
\begin{figure}
\begin{center}
% \fbox{\rule{0pt}{2in} \rule{.9\linewidth}{0pt}}
\includegraphics[width=1\linewidth]{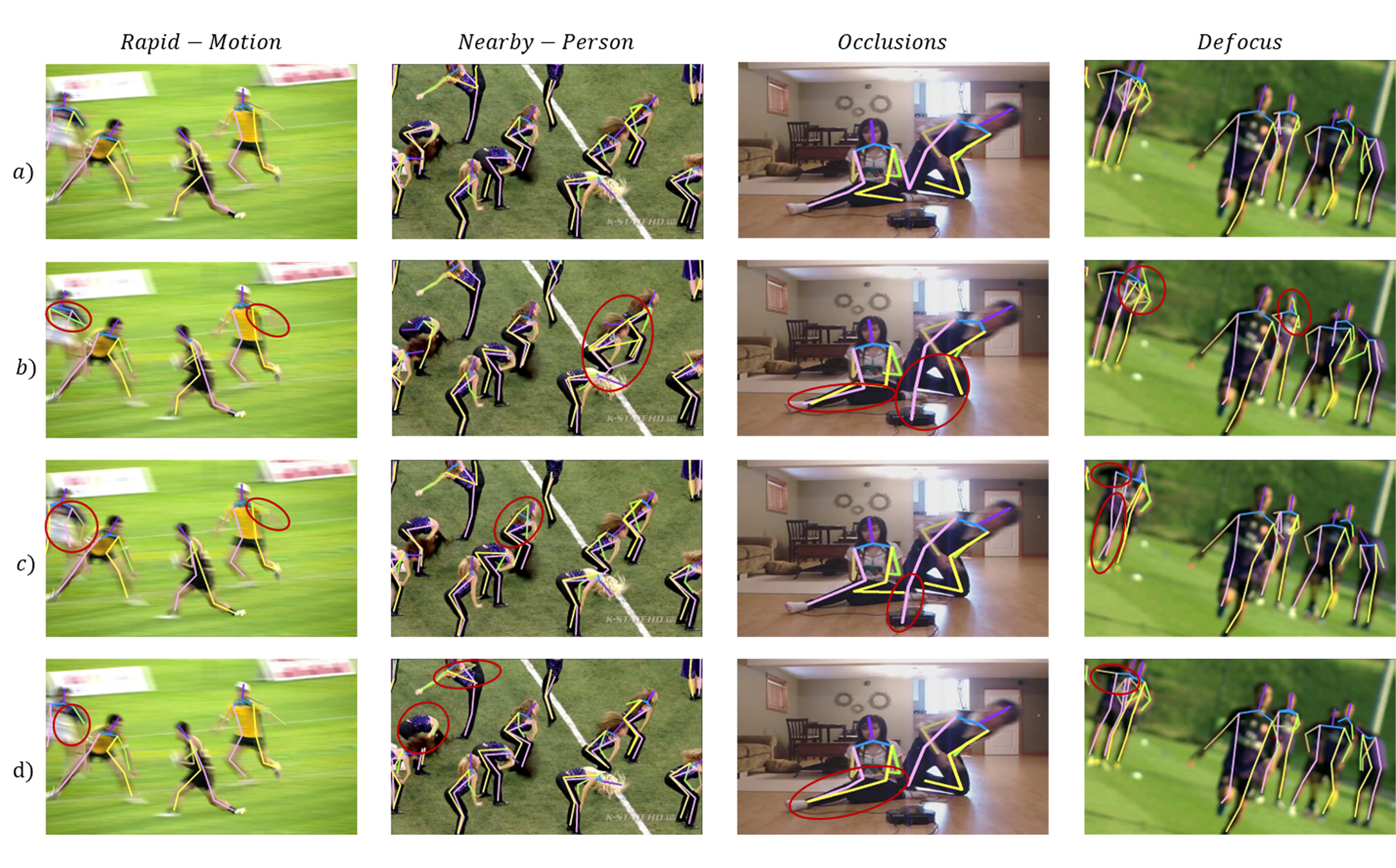}
\end{center}
\caption{Visualization of the pose predictions of our DCPose model($\bf a$), SimpleBaseline($\bf b$), HRNet($\bf c$), and PoseWarper($\bf d$) on the challenging cases from the $ PoseTrack 2017$ and $ PoseTrack 2018$ datasets.
Each column from left to right represents the Rapid-Motion, Nearby-Person, Occlusions, and Video-Defocus scene, respectively.
Inaccurate predictions are highlighted with the red solid circles.}
\label{fig:vis}
\vspace{-0.5em}
\end{figure}

\textbf{Pose Correction Network}\quad
We study the effects of adopting different sets of dilation rates for the convolutions in PCN. This corresponds to different effective receptive fields. We experiment with four different dilation settings: $d=3$, $d\in\{3, 6\}$, $d\in\{3, 6, 9\}$ and $d\in\{3, 6, 9, 12\}$ whereas the complete DCPose framework setting has $d\in\{3, 6, 9, 12, 15\}$. From the results in Table \ref{tab5}, we observe the gradual improvement of the \textbf{mAP} with increasing levels of dilation rates, from $81.7\to81.9\to82.7\to82.6\to82.8$. This is in line with our intuitions that increasing the depth and scope of the effective receptive fields, \emph{i.e.}, by varying the range of dilation rates, the Pose Correction Network is able to model local and global contexts more efficiently, leading to more accurate estimations of the joint locations.

% \renewcommand\arraystretch{1.2}
% \begin{table}
%   \resizebox{0.48\textwidth}{!}{
%   \begin{tabular}{c|c|c|c|c|c|c|c|c}
%     \hline
%   Time Span    &Head   &Shoulder &Elbow       &Wrist   &Hip    &Knee   &Ankle   &{\bf Mean}\cr
%     \hline
%   \textbf{$1$}&$\bf 88.4$&$\bf 88.8$&$\bf 84.1$&$\bf 78.4$&$\bf 83.0$&$\bf 81.5$&$\bf 74.3$&$\bf 83.0$\cr
%     $2$        &$87.7$ &$88.4$   &$83.8$      &$78.0$  &$83.0$ &$80.8$ &$73.2$  &{\bf $\bf 82.6$}\cr
%     $3$        &$87.6$ &$88.4$   &$83.6$      &$77.7$  &$82.6$ &$80.7$ &$71.7$  &{$\bf 82.1$}\cr
%     $4$        &$81.2$ &$87.4$   &$83.3$      &$76.7$  &$81.9$ &$80.2$ &$72.1$  &{$\bf 81.7$}\cr
%   %  $5$        &$83.8$ &$81.6$   &$77.1$      &$70.0$  &$77.4$ &$74.5$ &$70.8$  &{$\bf 77.0$}\cr
%     \hline
%     \end{tabular}}
%     \vspace{0.2em}\caption{\footnotesize{Ablation studies of different time span $T$.}}  \label{tab6}
% \end{table}

\textbf{Bidirectional Temporal Input}\quad
In addition, we examine the effects of removing a single temporal direction, \emph{i.e.}, removal of either ${F}_p$ or ${F}_n$ from the input to PTM and PRF. In removing the previous (respectively next) frame $F_p$ (respectively $F_n$), the \textbf{mAP} drops 0.6 (respectively 0.5). This highlights the importance of leveraging dual temporal directions as each direction allows access to useful information that are beneficial in improving pose estimation for videos.

\textbf{Time Interval $T$}\quad
The time interval $T$ described in Section \ref{section:method} is a hyper-parameter whose default value is set to $1$. In other words, DCPose looks at 3 consecutive frames as the shorter time/frame interval ensures the validity of our assumptions of temporal consistency. We experiment with enlarging the interval between frames, where $T$ is set to 2, 3, 4. Indices $p$ and $n$ are randomly selected with $p\in [c-T,c)$ and $n\in (c,c+T]$. The results reflect a performance drop with an increase in $T$, whereby the \textbf{mAP} decreases from 82.8 for $T=1$ to 82.6, 82.1, 81.7 as $T$ goes to 2, 3, 4 respectively. This is in accordance with our expectations, since frame-wise difference is small for shorter time intervals so that ${F}_p$ and ${F}_n$ can provide more abundant and effective temporal and spatial cues for our PTM and PRF. Conversely, a large time interval would invalidate our assumption of time consistency as the frame differences would be too great and the additional cues would lack efficacy. This ablation experiment explains our choice of modeling consecutive frames. The continuous frames of a video clip facilitate a precise aggregation of spatiotemporal context cues and improve the robustness of our model.

\subsection{Comparison of Visual Results}
In order to evaluate the capability of our model to adapt to sophisticated scenarios, we illustrate in Fig.~\ref{fig:vis} the side-by-side comparisons of our DCPose network against state-of-the-art approaches. Each column depicts different scenario complications including \emph{rapid motion, nearby-person, occlusions} and \emph{video defocus}, whereas each row displays the joint detection results from different methods. We compare our a) DCPose against 3 state-of-the-art methods, namely b) SimpleBaseline \cite{xiao2018simple}, c) HRNet-W48 \cite{sun2019deep} and d) PoseWarper \cite{bertasius2019learning}. It is observed that our method yields more robust detection for such challenging cases. SimpleBaseline and HRNet-W48 are trained on static images and fail to capture temporal dependency among video frames, resulting in suboptimal joint detection. On the other hand, PoseWarper leverages the temporal information between video frames to warp the initial pose heatmaps but only employing one adjacent video frame as the auxiliary frame. Our DCPose network makes full use of the temporal context by processing consecutive frames in dual directions. Through our principled design of PTM and PRF modules for better encoding of these information as well as PCN for refining pose detection, our method achieves new state-of-the-art both quantitatively and qualitatively.

\section{Conclusion}
In this paper, we propose a dual consecutive network for multi-frame person pose estimation which significantly outperforms existing state-of-the-art methods on the benchmark datasets. We design a Pose Temporal Merger and a Pose Residual Fusion module that allows abundant auxiliary information to be drawn from the adjacent frames, providing a localized and pose residual corrected search range for location keypoints. Our Pose Correction Network employs multiple effective receptive fields to refine pose estimation in this search range, achieving notable improvements and is able to handle complex scenes. %Future work includes reducing the quantization error between the heatmaps and coordinates.

\section{Acknowledgements}
This work was supported by the Natural Science Foundation of Zhejiang Province, China No. LQ19F020001, the National Natural Science Foundation of China No. 61902348,  the Humanities and Social Science Fund of the Ministry of Education of China under Grant No.17YJCZH076, and the Zhejiang Science and Technology Project under Grant No.LGF18F020001.

{\small
\bibliographystyle{ieee_fullname}
\bibliography{References}
}

\end{document}